\documentclass{article}

\usepackage[final,nonatbib]{neurips_2021}

\usepackage[utf8]{inputenc} 
\usepackage[T1]{fontenc}    
\usepackage{hyperref}       
\hypersetup{colorlinks,citecolor=red,urlcolor=blue,bookmarks=false,hypertexnames=true}
\usepackage{url}            
\usepackage{booktabs}       
\usepackage{amsfonts}       
\usepackage{nicefrac}       
\usepackage{microtype}      
\usepackage{xcolor}         
\usepackage{graphicx}
\usepackage[ruled, vlined]{algorithm2e}
\usepackage{ulem}           
\usepackage{amsmath}

\newif\ifturnoffcomments
\turnoffcommentsfalse

\title{Multi-task curriculum learning in a complex, visual, hard-exploration domain: Minecraft}

\author{%
   Ingmar Kanitscheider,\thanks{{\bf Contribution Statement}: Ingmar Kanitscheider, Joost Huizinga and Jeff Clune conceived the project, provided guidance and wrote the manuscript. Ingmar Kanitscheider designed and ran the experiments. David 
  Farhi, William Hebgen Guss, Brandon Houghton and Peter Zhokhov designed and tested the "Simon Says" task and the static exploration bonus. All authors worked on building out Minecraft as an RL environment.}$^{\;\; ,}$\thanks{{\bf Corresponding authors:} Ingmar Kanitscheider (\texttt{ingmar@openai.com}), Joost Huizinga (\texttt{joost@openai.com}), Jeff Clune (\texttt{jeffclune@openai.com})}\; 
  Joost Huizinga\setcounter{footnote}{1}\footnotemark,\; 
  David Farhi,\; 
  William Hebgen Guss,\\ 
  {\bf Brandon Houghton},\; 
  {\bf Raul Sampedro},\;
  {\bf Peter Zhokhov},\;
  {\bf Bowen Baker},\;
  {\bf Adrien Ecoffet},\\
  {\bf Jie Tang},\;
  {\bf Oleg Klimov},\;
  {\bf Jeff Clune}\setcounter{footnote}{1}\footnotemark \\ \\
  OpenAI

}

\begin{document}

\maketitle

\begin{abstract}
An important challenge in reinforcement learning is training agents that can solve a wide variety of tasks. If tasks depend on each other (e.g. needing to learn to walk before learning to run), curriculum learning can speed up learning by focusing on the next best task to learn. We explore curriculum learning in a complex, visual domain with many hard exploration challenges: Minecraft. We find that learning progress (defined as a change in success probability of a task) is a reliable measure of learnability for automatically constructing an effective curriculum. We introduce a learning-progress based curriculum and test it on a complex reinforcement learning problem (called “Simon Says”) where an agent is instructed to obtain a desired goal item. Many of the required skills depend on each other. 
Experiments demonstrate that: (1) a within-episode exploration bonus for obtaining new items improves performance, (2) dynamically adjusting this bonus across training such that it only applies to items the agent cannot reliably obtain yet further increases performance, (3) the learning-progress based curriculum elegantly follows the learning curve of the agent, and (4) when the learning-progress based curriculum is combined with the dynamic exploration bonus it learns much more efficiently and obtains far higher performance than uniform baselines. 
These results suggest that combining intra-episode and across-training exploration bonuses with learning progress creates a promising method for automated curriculum generation, which may substantially increase our ability to train more capable, generally intelligent agents.\setcounter{footnote}{0}\footnote{This is an experiment in sharing "partial work", meaning research that sheds light on a subject, but is not as complete as we would make it were we planning on publishing it as "complete work" in a peer-reviewed venue. Due to other priorities, we do not plan to do all that would be required for that level of scientific rigor. We thus faced a choice: either share it "as is", or not share it at all. We chose the former. We acknowledge much is missing, such as a more thorough literature review, experimental comparisons to other methods, ablations, etc. Nevertheless we believe that our results provide meaningful insights to the machine learning community. Our motivation is to share what we discovered, while minimizing the overhead in time and compute resources required to share the work.}

\end{abstract}

\section{Introduction}
An important challenge in reinforcement learning (RL) is to train agents that can solve a wide variety of tasks. Tasks often vary in difficulty and depend on each other, such that learning easier tasks first may help with learning more difficult tasks later. Just as human children crawl before they can
walk, and walk before they can run, interesting use cases for RL agents have difficulty hierarchies we would like to exploit: Learning first how to write syntactically correct code might help an agent to later learn how to fix semantic errors in a program. Learning first how to solve high school physics problems likely helps learn college-level physics. Such dependencies particularly exist in RL, where the problem of exploration
may prevent a policy from experiencing any useful gradient on a hard
task before it has mastered a prerequisite task. If the goal is to pass
a set of unit tests in a programming domain, an agent will not get any
useful feedback if it always fails all unit tests because it is unable
to write syntactically correct code. Side-stepping exploration by
collecting human demonstrations can be an option, but in many domains of
interest it might be difficult or impossible to collect enough
high-quality data to reach expert-human-level performance. Furthermore,
if the model relies purely on imitating humans its maximum performance
will be limited by the best demonstrations in our data set, and even the
combination of all the best demonstrations that humanity has to offer
probably will not move the model far beyond human performance.

In the following we assume that we're given a collection of RL tasks, 
each of which consists of a reward function (that defines the goal of
the task) and an environment distribution. A naive solution to learning
all tasks would be to attempt to learn all tasks simultaneously by
uniformly sampling tasks irrespective of the dependencies between them.
This solution ensures that the agent experiences a useful gradient as
long as some of the tasks are learnable given the agent's current skill
level. However, if the fraction of learnable tasks at any moment is very
small, it is computationally very inefficient, as the agent spends a lot
of time on tasks where it does not actually learn anything.

A more efficient alternative is to build a curriculum that narrows the
distribution of tasks being trained on to those that are currently
learnable. Which tasks are learnable at a given point in time, or in
what order tasks are most easily learnable, is typically not known in
advance. As such, the present paper explores methods for how we can infer
learnability on the fly and build an automatic curriculum over tasks. In
particular, our goal was to, given a set of tasks of varying difficulty,
learn as many tasks as fast and compute-efficient as possible.

This project focused specifically on tasks with identical environment
distributions, but different reward functions. Having tasks with
different goals but the same environment distribution is a natural
setting for the powerful models we wish to create, as the real world
presents a diverse but fixed environment distribution in which we would
want the model to perform many different tasks. The same is true for our
current generation of models, which generally have to learn many
different tasks (e.g. write different programs, summarize different
books, answer different questions, generate different images) in a
universal environment such as vision or language.

A curriculum only helps if mastering one task makes another task easier
to learn. However, in a goal-conditioned setup, even when tasks are
learned in the "correct" order, meaning that easier tasks are learned
before the harder tasks that depend on them, it can be difficult to
learn the harder tasks. One problem is that the agent does not know the
relationship between different tasks, meaning that if the agent is given
the goal for a task that it hasn't learned yet, it does not know which
of its previously learned behaviors might help it achieve that goal.
Another problem is that, even if the agent did know exactly which of the
previously learned tasks is related to the task it is currently trying
to solve, the behavior on that previously learned task may not
generalize to the current task (meaning that executing the behavior
learned on the previous task does not result in any zero-shot
performance on the current task), because the tasks are too different.
For example, even if the agent has learned to write syntactically
correct code and is executing that behavior frequently, it may never
write a program that passes a particular unit-test when that unit-test
is selected as the current task and thus be unable to learn to write a
program for it. We find that adding a goal-independent exploration bonus
that rewards all tasks the agent has not yet learned helps the agent
learn new tasks.

We have developed and evaluated these curriculum learning methods in a
Minecraft environment based on the MineRL platform\cite{guss2019minerl}. Minecraft is a well-thought-out visual world with rudimentary physics, which has the potential to allow our agents to learn many useful skills such as visual processing, spatial awareness, inferring causality and conducting experiments. Most relevant to curriculum learning in particular, Minecraft features a large tech-tree with many dependencies, making it relatively straightforward to define tasks of varying difficulty that depend on each other to varying degrees. In our experiment, in each task the agent is asked to obtain one out of 107 Minecraft items on command ("Simon Says"), some of which are deep into the tech tree (Figure \ref{fig:4}).

\begin{figure}[h!]
    \centering
    \includegraphics[width=\linewidth]{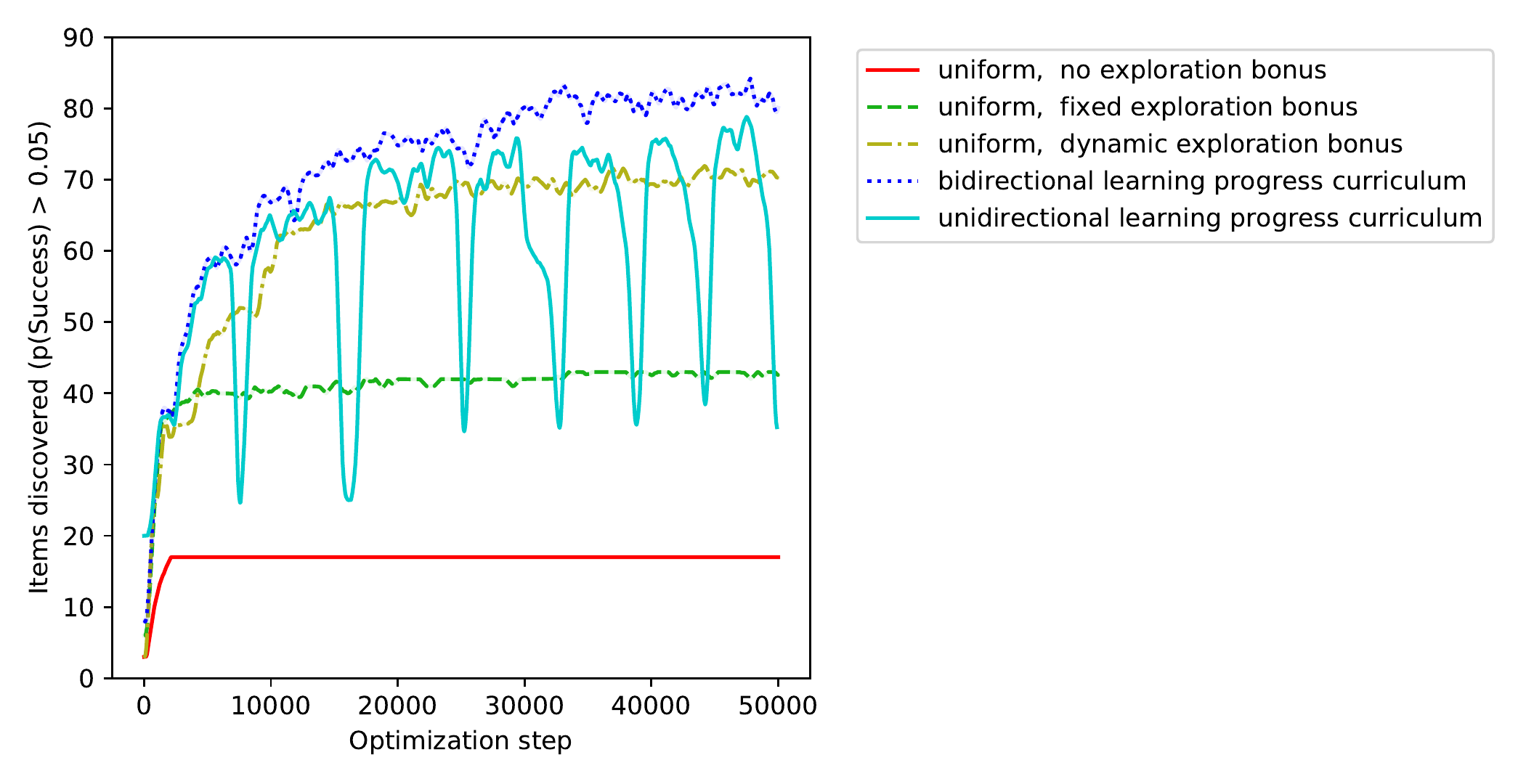}
    \caption{Number of discovered “Simon Says” target items (out of a total of 107 items) as a function of optimization steps for each of the training schemes we explore. Our goal was to maximize the number of items the agent is able to obtain and not to focus on obtaining high success probabilities; we therefore classify items with success probability larger than 5\% as discovered\protect\footnotemark. The bidirectional learning progress curriculum discovers the largest number of items, followed by uniform sampling with dynamic exploration bonus, uniform sampling with fixed exploration bonus and uniform sampling without any exploration bonus. The unidirectional learning progress curriculum at times discovers almost as many items as the bidirectional learning progress curriculum, but undergoes cycles where it forgets and rediscovers a large fraction of discovered items. We found inter-run variation to be low, and each run was expensive (21d on 32 GPUs), so we only plot one run per treatment.}
    \label{fig:1}
\end{figure}

Our key results are:
\begin{itemize}
    \item Uniform sampling does not learn many tasks, and learning flatlines at a low level (Figure \ref{fig:1}, red).
    \item An exploration bonus (as an auxiliary reward) to perform tasks unrelated to the current goal (a.k.a. curiosity search \cite{stanton2016curiosity, stanton2018deep}) substantially improves performance (Figure \ref{fig:1}, green).
    \item Adding an additional across-training diversity pressure (similar to novelty search \cite{lehman2011abandoning} and intrinsic motivation \cite{strehl2008analysis}) by removing the exploration bonus dynamically for items the agent can already reliably obtain further improves performance (Figure \ref{fig:1}, yellow).
    \item Adding a learning progress curriculum increases performance even more (Figure \ref{fig:1}, dark dotted blue). A video of a successful agent obtaining challenging items high up in the tech tree can be viewed at \href{https://youtu.be/MFDudOvn3oc}{https://youtu.be/MFDudOvn3oc}.
    \item With the learning progress curriculum, the sampling of tasks elegantly follows the learning curve of the agent, focusing learning on the frontier of the agent's skill set as that skill set expands (Figure \ref{fig:3}, bidirectional learning-progress curriculum).
    \item Looking for \textit{any} learning change (including performance \textit{drops}) (Figure \ref{fig:1}, dark dotted blue) prevents the catastrophic forgetting of previously learned tasks that otherwise occurs when you only measure learning \textit{improvements} (Figure \ref{fig:1}, light solid blue), hurting overall performance.
\end{itemize}

\footnotetext{We found that the 5\% threshold clearly shows the differences between the different treatments within 50,000 optimizers steps, but we expect that similar results could be obtained for higher success-probability thresholds given sufficient additional training.}

\begin{figure}[h]
    \centering
    \includegraphics[width=\linewidth]{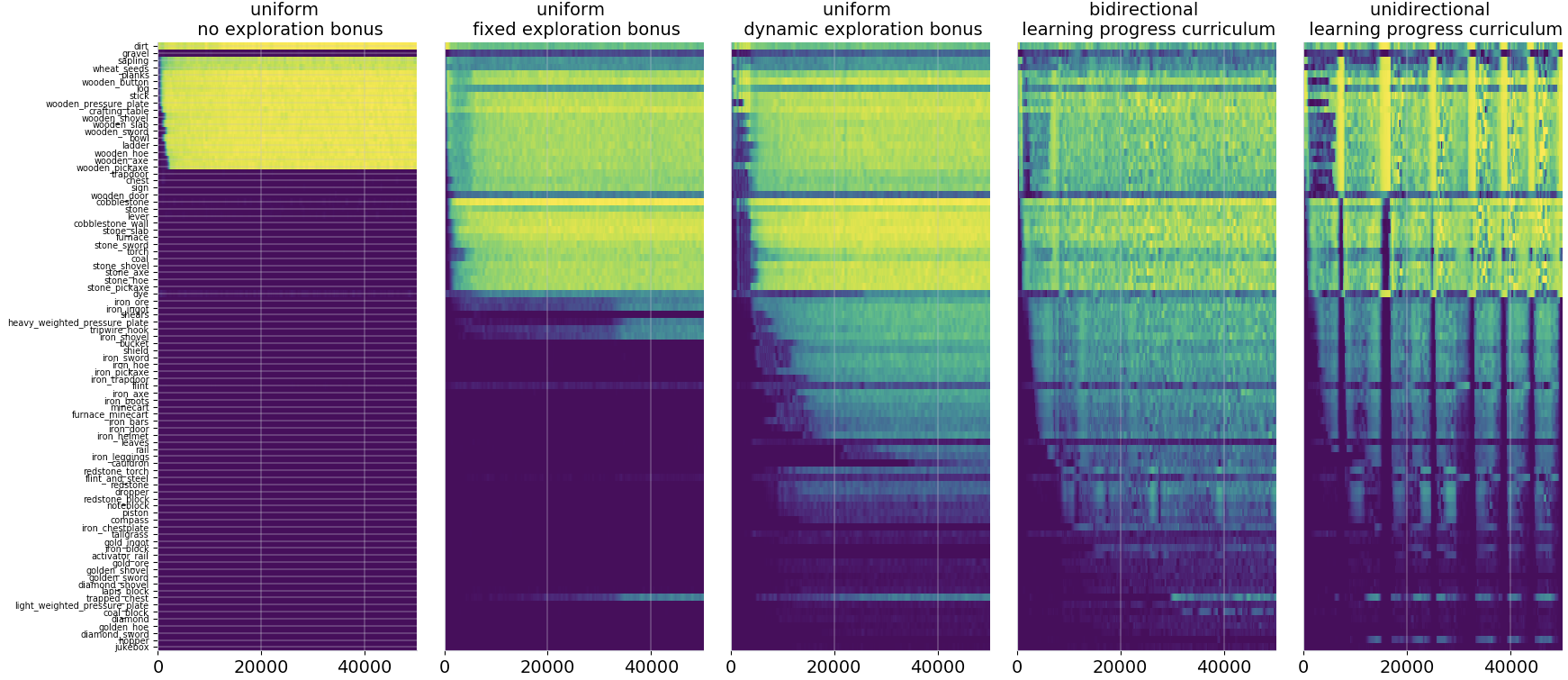}
    \caption{Conditional success probabilities of individual "Simon
Says" target items under each treatment. Items are ordered according to
the order in which the bidirectional learning-progress curriculum learns
to collect the items. We only show items that are discovered by at least
one of the treatments. The bidirectional learning progress curriculum
discovers a super set of items discovered by other treatments.}
    \label{fig:2}
\end{figure}

\begin{figure}[h]
    \centering
    \includegraphics[width=\linewidth]{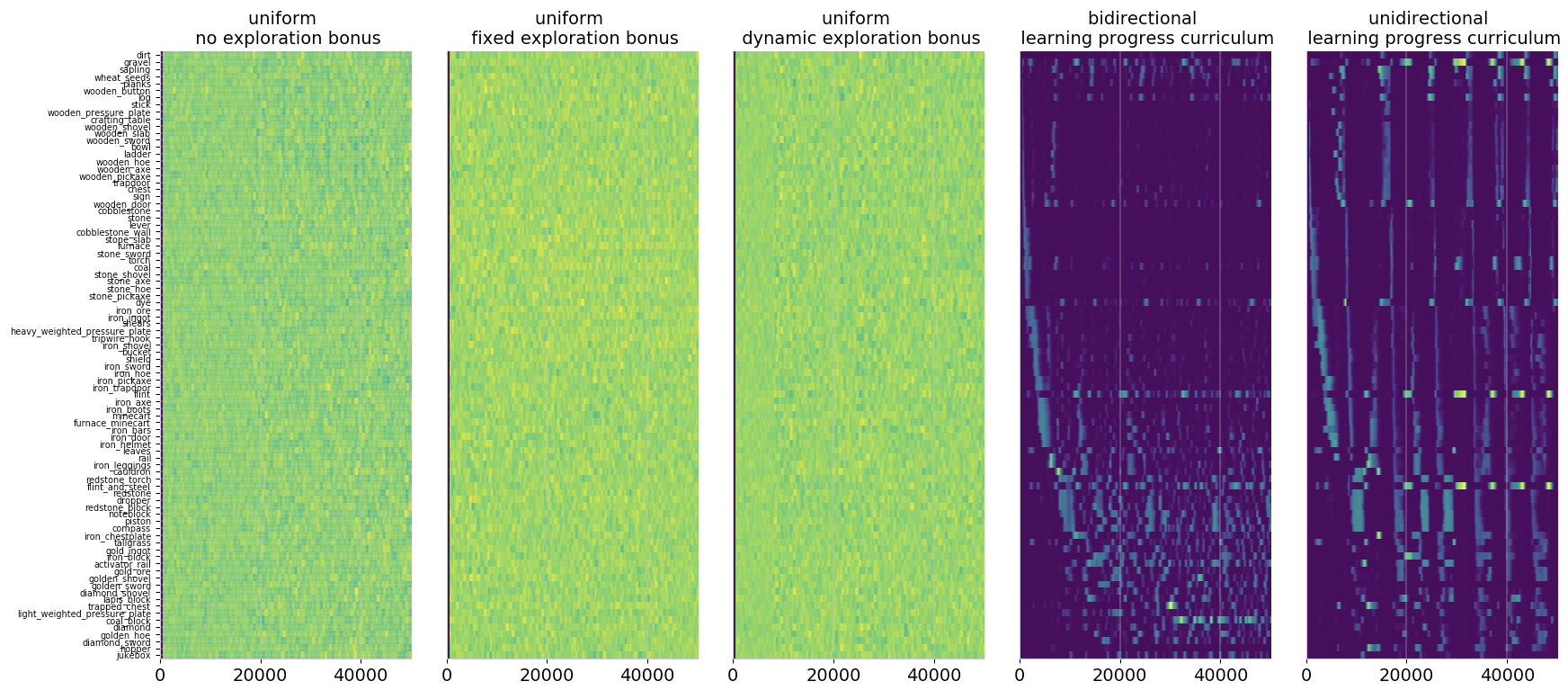}
    \caption{Probability of sampling individual “Simon Says” target items under each treatment. Items are ordered as in Figure
\ref{fig:2}. The learning progress curricula accurately track,
and thus sample, items whose success probability changes (for
bidirectional-learning progress) or increases (for
unidirectional-learning progress) the most. }
    \label{fig:3}
\end{figure}

\begin{figure}[h]
    \centering
    \includegraphics[width=\linewidth]{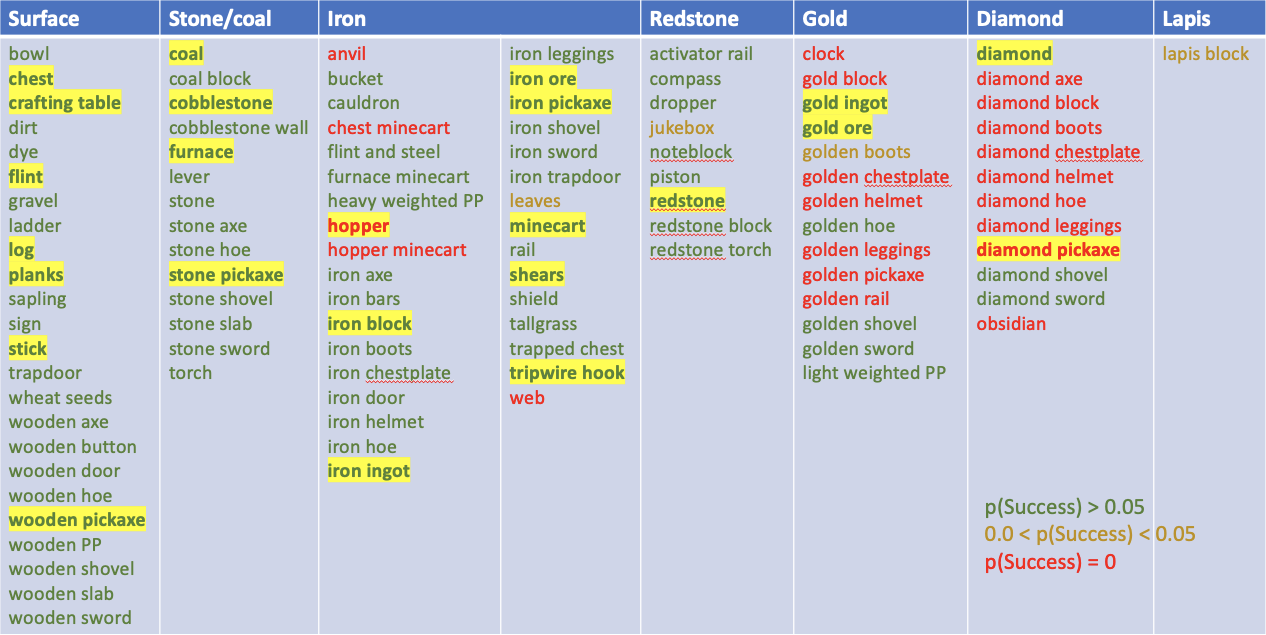}
    \caption{Set of 107 "Simon Says" Minecraft items that the
agent is tasked to obtain. The left-most column contains items the agent
can obtain at the surface without mining. The remaining items are
grouped by the mining resource ("stone", "coal", "iron", "lapis",
"redstone", "gold", "diamond") that is required to craft them. From left
to right, item categories are roughly ordered by how difficult they are
to obtain. Difficulty is mainly determined by how deep they are in the
Minecraft tech tree and by how rare they are to find. The highlighted
items are prerequisite items for downstream items, and thus represent
essential stepping stones in the curriculum. The color code indicates
the final success probability of our best treatment (the bidirectional
learning progress curriculum, presented below). The agent obtains 82
items with success probability > 0.05, 4 items have non-zero success
probability below 0.05, and 27 items have zero success probability. In
comparison, a naive baseline (uniform sampling only) obtains only 17
items with success probability > 0.05.}
    \label{fig:4}
\end{figure}

\section{Methods}

\subsection{Learning progress curriculum}

First, we identify the conditions where we expect curriculum learning to
work much better than uniform sampling: assume we are given a long list
of mutually dependent and progressively more difficult RL tasks $T_1,...,T_N$. We further assume that a policy that has mastered $T_i$ can learn $T_{i+1}$, but not $T_{i+2}$. An agent can therefore learn all
tasks if they are presented in this order. However, learning all tasks
under uniform sampling is much harder, because only every $1/N$-th rollout
is sampled from a learnable task. At a minimum, if it takes at least $H$
samples per batch from task $T$ to learn it, we will need to collect $N$
times more samples (i.e. train with an $N$-times larger batch size) when
performing uniform sampling just to ensure that we get $H$ samples from
task $T$ per batch. In addition, the non-learnable tasks may add
additional noise to each batch without providing any signal, meaning an
even larger than $N$-times larger batch size may be needed. The optimal
curriculum is therefore much more compute efficient than uniform
sampling.

A key requirement of designing an automatic curriculum is to infer which
new task may be learnable given the current skill \textit{a priori}. A common
approach is to use success probability as a proxy for learnability.
Typically, one defines a lower success probability threshold below which
tasks are considered too hard and an upper success probability threshold
above which tasks are considered too easy and one would predominantly
sample tasks between these two thresholds \cite{wang2019paired, wang2020enhanced, openAI2019solving}. However,
this approach has several shortcomings: First, if the initial state of
the environment is randomized, the agent may have intermediate success
probability on a task solely because the task is too easy from some
initial states and too hard (or even impossible) from others, thus
preventing the agent from improving any further. In Minecraft, for
example, it is only possible to collect jungle planks when starting in a
jungle biome. If the agent only starts in the jungle 20\% of the time,
its success probability for jungle planks would be capped at 20\%, even
if it perfectly learned the task. Second, success probability thresholds
that correlate well with learnability might in general depend on the
task, the initial state of the environment and the learning stage of the
agent. Finally, stochastic environments (i.e. environments with a
stochastic transition function) can have a similar disconnect between
success probability and learnability as environments with random initial
states: An easy-to-learn task may be capped at an intermediate success
probability because of an uncontrollable stochastic event that prevents
the agent from succeeding reliably. Selecting tasks with intermediate
success probability might therefore select tasks where the agent cannot
learn anything new.

Instead, we infer learnability of tasks by measuring learning progress,
i.e. the recent change in success probability for each task. Given such
a measure, the curriculum predominantly samples tasks with large
learning progress. We explore sampling based on a bidirectional learning
progress measure (that tracks both increases and decreases in success
probability) and a unidirectional measure (that only tracks increases in
success probability). The advantage of sampling based on the
bidirectional measure is that it not only samples novel tasks when they
start showing learning progress, but also samples tasks that are being
forgotten.

\subsection{Learning progress inference}

\begin{figure}[h]
    \centering
    \includegraphics[width=\linewidth]{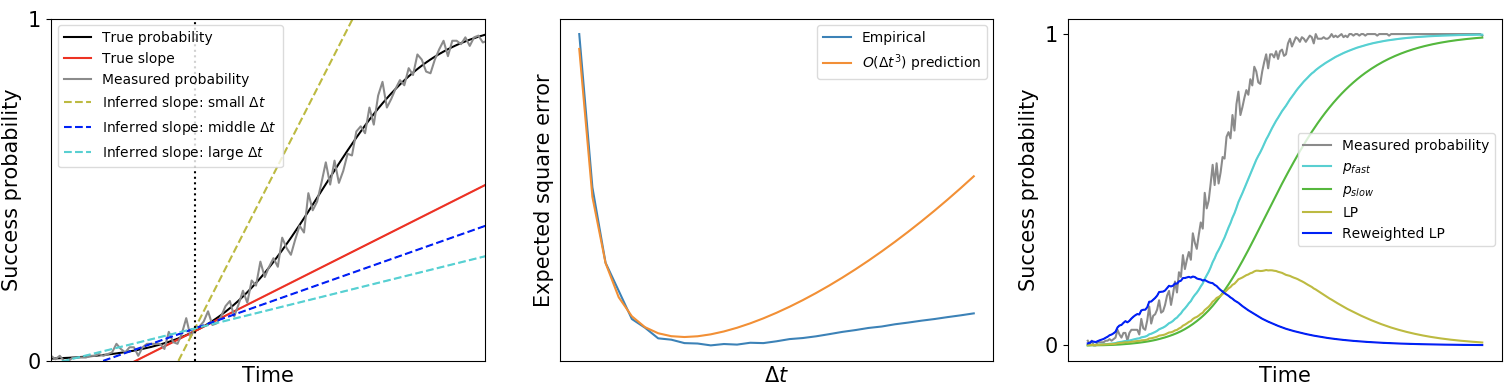}
    \caption{Inference of learning progress on an example
fictional problem from the slope of the measured success probability
curve.}
    \label{fig:5}
\end{figure}

When designing a process to infer learning progress from empirical
measurements of successes and failures on a given task it is important
to consider the parameters that influence the accuracy of a learning
progress estimator. In particular, as learning progress is measured in
terms of how the success probability changes over time, it is important
to pick the appropriate time scale $\Delta t$ over which the
before/after change in success probability is measured. Consider the
following example of a task whose true success probability (black)
increases over time (Figure \ref{fig:5}, left).

We would like to infer the true learning progress at $t=0$ (vertical dotted
black line), which we define as the slope of the success probability
curve (solid red line), from the success probabilities measured at time
snapshots $t$ and $t - \Delta t$. However, we only have
access to a history (up to $t$) of noisy measurements of success
probability (jagged grey line; the measurements here are sampled from a
binomial distribution with the success probability $p$ equal to the value
of the true probability - the black line - at time $t$ and $n$, the
number of samples, set to 200): If we choose $\Delta t$ too
small, we end up measuring the slope of the noise (dashed yellow line)
instead of the slope of the underlying trend (with different noise the
slope of yellow could have actually been sharply negative!); the
variance of our estimator is too high. If we choose $\Delta t$
too large we do not fully capture recent changes in the slope (light
blue); our estimator is biased (in the statistical sense: the error is
non-zero even with infinite data). An intermediate $\Delta t$
gives us a good trade-off between bias and variance (dashed dark blue
line). Indeed, resampling from the binomial many times (each sample
being noisy owing to the small sample size) shows that the expected
square error is minimal for intermediate values of $\Delta t$
(Figure \ref{fig:5}, middle, blue) and we can show analytically that
this is in general true for any curved success probability curve (Figure
\ref{fig:5}, middle, orange and Appendix \ref{app:a}).

Besides picking the right time scale, we can improve the reliability of
our learning progress estimator by calculating it based on the success
probability measurements of all previous time steps, rather than just
two snapshots in time. We can implement this by replacing the snapshots
with two exponential mean average (EMA) processes with different time
scales. In particular, the first EMA represents a "fast success
probability" $p_{\text{fast}}$ estimate for each task (Figure
5, right, red). We then smooth $p_{\text{fast}}$ with a
second, identical EMA to obtain a "slow success probability" estimate
$p_{\text{slow}}$ (green). From there, we define learning progress as
the difference between the fast and slow success probability estimates
(Figure \ref{fig:5}, right, yellow). This learning progress estimate
increases after the success probability curves upward (meaning the agent
is going from no learning progress to increasing learning progress),
flattens after the success probability reaches a linear upward trend
(meaning learning progress is happening at a steady pace), and finally
goes down to zero again after the measured success probability converges
(meaning no further learning progress is happening because the agent has
learned everything there is to learn on this task). Note that learning
progress does not perfectly follow the derivative of the success
probability, but is delayed because it is calculated from two EMA
processes which themselves are delayed estimates of the success
probability. In practice, we are able to correct for this temporal
delay, as explained in detail below. Based on this definition of
learning progress, we further define bidirectional and unidirectional
learning progress as the absolute and rectified difference,
respectively, between fast and slow success probability. If
$p_{\text{fast}}$ is larger than $p_{\text{slow}}$, as in Figure
\ref{fig:5}, right, the two measures are identical.

We can improve the curriculum further by putting additional focus on
tasks that have not only high learning progress, but also a low success
probability. The rationale is as follows: Our goal is to learn as many
tasks as possible, as opposed to learning a few tasks very well. We thus
care much more about an improvement from 0.01\% to 5\% in a new task than
we do about moving a task we already perform well on from 95\% to 99.99\%
reliability (this is the same reason why we chose a 5\% threshold for the purpose of evaluation). Since low success
probability tasks are more likely to be novel, putting particular focus
on them helps to expand the set of learned tasks. In addition, low
success probability tasks are also harder to learn because the agent
observes fewer successes. Sampling hard tasks more often increases the
number of successes and therefore helps to mitigate the exploration
problem. While our focus on low-probability tasks may seem reminiscent of the fixed success-probability-threshold methods discussed above, it avoids the false positives when selecting tasks solely based on success thresholds because it excludes tasks without learning progress.

We implement this focus by applying a reweighting function to
$p_{\text{fast}}$ and $p_{\text{slow}}$ before computing learning
progress. The reweighting function we use magnifies learning progress in
tasks with low success probability at the expense of learning progress
in tasks with high success probability. However, tasks without learning
progress are still mapped to zero learning progress, no matter the
success probability (see Appendix \ref{app:b} for details). In the fictional
example above, reweighted learning progress is given by the blue curve
in Figure \ref{fig:5}, right. The figure illustrates that this reweighting can
also compensate for the temporal delay that we observed with the regular
learning progress estimate because success probabilities will be lowest
right when the agent starts to learn a particular task. Reweighting is
applied in the same way for bidirectional and unidirectional learning
progress.

As a last step, we use a sampling function that focuses 90\% of sampling
weight on roughly 20\% of tasks with the largest reweighted learning
progress. This sampling function was designed to strike a balance
between focusing only on the tasks with the highest learning
probability, which can cause catastrophic forgetting of all other tasks,
and uniformly sampling all tasks, which would waste a lot of computation
on tasks where no learning progress is observed (see Appendix \ref{app:c}).

In conclusion, by taking the difference between a fast and slow estimate
of the measured success probability we obtain an accurate, but delayed,
estimate of learning progress (Figure \ref{fig:5}, yellow). By reweighing
this estimate towards tasks with low success probability we compensate
for the delay and put the majority of focus on exactly those tasks that
are currently learnable, but have not been learned yet (Figure
\ref{fig:5}, blue). Taking 90\% of our samples from the tasks that score
within the top 20\% of this reweighed learning-progress metric thus
ensures that our curriculum always pushes on the frontier of the
currently learnable tasks that have not been learned (well) yet.

\subsection{Curricula over goal-conditioned tasks}

A key requirement of curriculum learning (at least using current RL
methods) is that the agent achieves a small success probability (within
available/reasonable compute) on a new task after mastering a previous
task. However, if tasks differ by goals that are given as observations
to the agent, the agent may not know how to interpret the new goal
observation and may just display random behavior instead of the behavior
of a prerequisite task. In addition, obtaining useful feedback in a
goal-conditioned setup is much harder than learning unconditional tasks
because the agent only experiences a positive reward for completing the
new task in episodes where the goal matches the task. The success
probability of the new task is therefore suppressed by the background
probability of sampling the corresponding goal, which makes the task
difficult to learn.

We can encourage the discovery of new goals by combining the
goal-conditioned main reward with a goal-independent exploration bonus
for all of the curriculum tasks, even if they are unrelated to the
current goal. This exploration bonus helps the agent to explore the
environment when given an unknown goal, thus increasing the chances of
success. In our Minecraft experiment, where a new goal corresponds to
collecting a new item, for each item in the set of potential goal items,
we provide a reward the first few times the agent collects that item in
an episode, regardless of the currently selected goal. Specifically, the
agent receives a reward of $0.5^N$ for the $N$-th collection of the
same item, i.e. the reward decays with a factor of 0.5 for each
subsequent collection. In addition, the agent only receives a reward for
a subsequent collection of item $X$ if $N$ is larger than the number of
items $X$ in its inventory at any previous time during the episode (this
constraint prevents the agent from just racking up reward by dropping
and picking up items). The exploration bonus therefore incentivizes the
agent to constantly collect or craft new items that it hasn't held
previously during the episode. This idea of encouraging an agent to do
as many new, different things within an episode is similar to previous
work in Curiosity Search \cite{stanton2016curiosity, stanton2018deep}.

The exploration bonus is combined with the main reward by simply summing
the two rewards, meaning they have to be carefully balanced: the agent
may favor the exploration bonus over the main reward if the exploration
bonus is too high or not gain any benefits from the exploration bonus
when it is too low. In addition, this balance will have to change as the
curriculum advances; at the start of training it is fine for the agent
to get an exploration bonus for all of the easy tasks, but as the
curriculum moves towards harder tasks it becomes increasingly likely
that the agent will spend most of the limited time it has per episode
collecting reward by obtaining the exploration bonus for all easy items
rather than attempting to complete the hard task selected by the
curriculum. In our experiments, this balance is maintained across
training by a second curriculum that successively removes
already-learned goals from the exploration bonus. We call this
curriculum-based exploration bonus the "dynamic exploration bonus".

In our Minecraft experiment, we implement the dynamic exploration bonus
in the following way: Using exponential mean averaging (EMA), we keep
track of the individual success probabilities in the main,
goal-conditioned, task. Only items for which the agent has a success
probability smaller than 0.1 are included in the set of items rewarded
by the exploration bonus (called "exploration set"). Thus, as the agent
improves its performance over training, items are successively removed
from the exploration set, but are added back if the performance on the
corresponding task drops again below 0.1. The dynamic exploration reward
can be seen as an implementation of across-training diversity pressure
similar to previous work in Novelty Search \cite{lehman2011abandoning} and Intrinsic
Motivation \cite{strehl2008analysis}.

\subsection{"Simon Says" task}

We study curriculum learning on a set of goal-conditioned Minecraft
tasks, in which the agent is tasked to collect one out of a set of 107
items from the Minecraft tech tree (Figure \ref{fig:4})\footnote{In addition to the 107 items displayed in Figure \ref{fig:4}, the target set also contains 6 additional items that we later realized were impossible for the agent to obtain.}. Some items (such as "dirt") can be very easily
obtained, while other items (such as "diamond") are rare and also
require first obtaining many other resources and crafting required
tools. The agent observes the target item as a one-hot encoding. It has
5 minutes (1500 time steps) to complete the task and obtains a reward of
+1 upon success. After each success or failure a new task is selected
without resetting the world or respawning the agent. Tasks may be
sampled uniformly or from a distribution that is determined by a
curriculum.

The maximum episode length is 30 minutes (9000 time steps), but the
episode is terminated prematurely if the agent has failed two
consecutive tasks or dies of other causes, such as drowning, falling off
a cliff or falling in lava. After an episode ends, a new episode is
started in a new environment, but the agent has a 50\% chance to retain
the inventory from the end of the previous episode. In preliminary
experiments, we found that this "inventory inheritance" leads to
slightly faster learning, as the agent does not always have to gather
all the necessary prerequisite items from scratch when trying to learn
how to obtain difficult items deep in the tech tree. Because each run was computationally expensive (21d on 32 GPUs) we only plot one run per treatment, but we found inter-run variation to be low.

\section{Results: Evaluation on Minecraft "Simon Says"}
\subsection{Uniform sampling without exploration bonus}

In the standard Simon Says task, the agent only experiences a positive
reward for obtaining an item if the item corresponds to the goal of the
task. This makes all but the easiest tasks difficult to learn, because
the success probability of the task is suppressed by the probability of
sampling the corresponding goal and because there is no other
exploration bonus. As expected, the results with this method are poor:
the agent only learns to obtain 17 out of 107 tasks (Figure \ref{fig:1},
red line). Note that we say that the agent has "learned" a task if it
achieves a success probability of at least 5\%. Worse, the plot shows
that learning plateaued early and does not improve over time. The agent
only discovers a subset of items that can be obtained on the surface,
such as "dirt", "sapling" and a number of wooden tools (Figure
\ref{fig:2}, 1st from left).

\subsection{Uniform sampling with fixed exploration bonus}

To support exploration of harder tasks, we add the exploration bonus
over all items in the Simon Says 107 set. The exploration bonus is added
to the main Simon Says reward with a coefficient that was tuned to each
condition. For uniform sampling without curriculum we found in
preliminary experiments that a coefficient of 0.05 performs best.

Adding the exploration bonus increases the number of explored items at
convergence from 17 to 43 (see Figure \ref{fig:1}, dotted green line).
The agent discovers almost all surface items, learns to mine
underground, learns to craft stone tools, and even discovers how to
create a few iron tools such as "iron ingot", "heavy weighted pressure
plate", "tripwire hook" and "iron shovel" (Figure \ref{fig:2},
2nd from left).

\subsection{Uniform sampling with dynamic exploration bonus}

However, while the exploration bonus helps the agent in solving a larger
number of tasks, it actually can make it hard to learn to collect harder
items that are deeper in the tech tree. The reason is that the
exploration bonus equally rewards items that the agent already knows how
to obtain (but has not yet obtained this episode) and items that the
agent still needs to learn how to get. When given a hard-to-obtain goal
that the agent has not learned yet, it may be more rewarding to collect
easy-to-obtain and previously learned items in order to collect their
exploration bonus, therefore "distracting" the agent from the task it is
currently supposed to complete. One straightforward solution to this problem is to provide an exploration bonus only for those items that the agent does not yet reliably know how to obtain.  This allows us to include the exploration
bonus only when it is useful, namely for learning how to obtain new
items, without distracting the agent.

The "dynamic exploration bonus" implements exactly this idea by removing
items whose goal-conditioned success probability in the main Simon-Says
task is larger than 0.1.

As we only give an exploration bonus for hard items that the agent
rarely gets, we can increase the coefficient of the exploration bonus
without increasing the extent to which the agent gets distracted by easy
items. In preliminary experiments we found that a coefficient of 0.5
performed best.

The dynamic exploration bonus further increases the number of explored
items at convergence to about 70 (see Figure \ref{fig:1}, dashed yellow
line). From within the target set, the agent discovers all surface
items, all stone items, most iron items and most redstone items (Figure \ref{fig:2}, 3rd from left).

Conceptually, the dynamic exploration bonus and the conditional Simon
Says reward interleave in the following way: At first, the exploration
bonus incentivizes the agent to learn how to obtain an item
unconditionally, i.e. irrespective of the current Simon Says task. As
the unconditional success probability increases, so does the conditional
success probability, that is, the success probability when the new item
corresponds to the goal. Once the conditional success probability
surpasses 0.1, the item is removed from the exploration set and is only
rewarded through the main Simon Says reward (i.e. only when given the
task where the item corresponds to the goal). The agent can then
solidify the association between the observed Simon Says task label and
the task.

\subsection{Learning progress curriculum}
\label{sec:lp_curr}

In this treatment we sample Simon Says tasks using the learning progress
curriculum instead of uniform sampling. As in the previous section, we
remove easy items from the exploration set using the dynamic exploration
bonus and we again set the overall coefficient of the exploration bonus
to 0.5. The learning progress curriculum improves task learning by two
mechanisms: First, by focusing on Simon Says tasks with the largest
learning progress it supports learning the conditional association
between task label and task. Second, the curriculum interacts with the
exploration bonus to facilitate the unconditional discovery of novel
items. If learning progress is made on an item such that the curriculum
frequently samples it, the agent is likely to obtain it more often and,
if that item is a prerequisite for any dependent items, the agent is now
more frequently in a position where it actually has this prerequisite
item and is thus able to collect the dependent items.

We tested both the bidirectional and unidirectional versions of the
learning progress curriculum (described above). With the bidirectional
learning progress curriculum, the agent discovers 82 items (see Figure \ref{fig:1}, dotted blue line). The agent discovers all surface items,
all stone items, most iron items, most redstone items, half of the
golden items and a few diamond items (Figure \ref{fig:2}, 2nd
from right). With the unidirectional learning progress curriculum, the agent discovers 79 items (see Figure \ref{fig:1}, solid light blue line and Figure \ref{fig:2}, 1st
from right), which is almost as many as the bidirectional learning progress treatment, but training in the unidirectional treatment is unstable because of many cycles of forgetting and rediscovery (see next section). Both curricula accurately sample items for which success
probability changes or increases the most (compare Figure \ref{fig:2}, 1st and 2nd from right with Figure \ref{fig:3}, 1st and 2nd from right).

With both the bidirectional and unidirectional learning-progress
curricula, the interaction between the dynamic exploration bonus and the
conditional Simon Says reward is similar to the interaction between the
dynamic exploration bonus and uniform sampling. However, with the
learning progress curricula the learning of the conditional association
(performing the task when asked) is more focused and accelerated,
because the learning progress curriculum detects the increase in
conditional success probability and consequently focuses on that task,
which means there will be many more rollouts where the task is also the
current goal and thus many more positive examples from which the agent
can learn the association.

\subsection{Mitigating catastrophic forgetting by tracking bidirectional learning progress}

A curious phenomenon of the unidirectional learning progress curriculum
is that it goes through cycles where the number of discovered items drop
sharply, before it again recovers. These cycles are caused by
catastrophic forgetting owing to correlations in learning progress
between groups of items. As an example, let us consider a case where,
after having discovered underground items, the agent improves its
success probability for a surface item such as "sapling". The
unidirectional learning progress curriculum samples "sapling" more often
which causes the agent to spend more time at the surface, which in turn
causes the agent to also improve its success probability for other
surface items (they are easier because the agent is already on the
surface), creating a positive feedback loop. Meanwhile, success
probabilities for underground items decrease because the agent spends
less time underground and thus forgets (catastrophically) how to perform
such tasks. The bidirectional learning progress curriculum would
immediately increase its sampling of underground items in order to
prevent those tasks from being forgotten, which would prevent
catastrophic forgetting and thus cycles from appearing. In contrast, the
unidirectional learning progress curriculum does not increase the
sampling of underground items when their success probabilities are
decreasing. As a consequence, it enters a period where it exclusively
focuses on surface items and generally these periods last long enough
for the agent to almost completely forget how to obtain underground
items when asked. Since only 24 out of the 107 potential goals are
surface items, this causes a large drop in the number of discovered
items. However, after about 1000-2000 optimizers steps, the curriculum
notices positive learning progress on the underground items again,
allowing it to rediscover underground items, and the number of
discovered items recovers (Figure \ref{fig:1}, solid light blue line).
Interestingly, much of the ability to perform these skills is still
latent in the network, as its performance recovery is swift.

Neural networks in general and RL agents in particular are subject to
catastrophic forgetting if the data distribution changes during
learning. The easiest method to mitigate catastrophic forgetting is to
sample the data i.i.d. (i.e. uniform sampling of discrete tasks),
whereas a curriculum might cause the agent to forget early tasks.

The success of the bidirectional learning progress curriculum shows that
monitoring previous tasks and resampling them if their performance drops
can be a powerful method to mitigate catastrophic forgetting. As shown
in Figure \ref{fig:3}, 2nd from right, only sporadic resampling
of old tasks is sufficient, which is much more compute efficient and has
better scaling properties than iid sampling of all previously learned
tasks.

\section{Related work}

\textbf{Automated curriculum learning} There is an extensive literature on
training neural networks with a curriculum \cite{elman1993learning}, see \cite{bengio2009curriculum} for an overview.
More recently, automated curricula have been
studied extensively in the context of RL in \cite{sukhbaatar2018learning, florensa2018automatic, matiisen2019teacher, portelas2020teacher, wang2019paired, wang2020enhanced, openAI2019solving, zhang2020automatic, campero2020learning, dennis2020emergent, gur2021adversarial, openai2021asymmetric}. Generally, tasks are selected
based on success probability or reward thresholds \cite{wang2019paired, wang2020enhanced, openAI2019solving, campero2020learning} or regret \cite{dennis2020emergent, gur2021adversarial}. Static-threshold-based methods
present an intuitive starting point for a curriculum, but have the
downside that they are difficult or even impossible to tune, as
discussed previously (Sec. \ref{sec:lp_curr}). Regret-based
methods calculate per-task regret by subtracting the average return over
several rollouts on that task from the maximum (known) return on that
task, and then preferrably select tasks where regret is high, with the
theory being that there is still a lot to learn on tasks where regret is
high \cite{dennis2020emergent, gur2021adversarial}. In the presence of
environmental stochasticity, this scheme may select more stochastic,
less learnable tasks at the expense of less stochastic, more learnable
tasks, because the maximum return may have been obtained under
particularly lucky conditions such that the average return will always
be much lower, despite there being nothing left to learn.
Learning-progress-based curricula, like the method in this paper, have
the potential to address these issues, as discussed next.

\textbf{Learning progress} was first proposed as a curiosity signal that
incentivizes agents to explore novel states in stochastic environments \cite{schmidhuber1991curious, oudeyer2007intrinsic, kim2020active}. Later, in \cite{graves2017automated, matiisen2019teacher, portelas2020teacher} learning progress was used as a measure to select
tasks in a curriculum. The novel contributions of our work are to
systematically study how learning progress can be measured reliably, to
apply learning progress curricula on hard problems that require RL at
scale, and to show how learning progress curricula over goals can be
combined with a dynamic, goal-independent exploration bonus.

\textbf{Curiosity Search} The static exploration bonus we examined
incentivizes the agent to obtain items that it has not obtained in the
current episode, and is thus a method for encouraging within-episode
exploration. Within episode exploration has previously been explored in
the Curiosity Search work by \cite{stanton2016curiosity, stanton2018deep}, who
demonstrated that it effectively encourages agents to explore their
environment, though they also demonstrated the downside of having to
explore the entire environment in every episode when trying to perform
deep exploration.

\textbf{Intrinsic motivation} The dynamic exploration bonus, on the other
hand, changes over training and encourages the agent to obtain different
items, not within a single episode, but across episodes. Exploration
across episodes has been extensively studied in the form of count-based
exploration (e.g. \cite{strehl2008analysis}), where the algorithm tracks
the number of times each state has been visited over training and
provides a bonus reward to the agent for visiting each state that is
inverse proportional to the number of times that state has been visited.
\cite{bellemare2016unifying} adapted count-based exploration to work in
state-spaces that are too large to keep individual counts, and they
demonstrated some success in deeply exploring sparse reward
environments. However, later work \cite{ecoffet2019go, ecoffet2021first} hypothesized
that unconditional count-based exploration methods can suffer from
detachment, where the agent consumes all the intrinsic reward towards
one or more states and forgets how to return to those states, and
derailment, where the exploratory mechanism of the algorithm can prevent
the agent from returning to a previously visited state. Our
bidirectional learning-progress curriculum avoids detachment by
immediately increasing the sampling rate of any goal where success
probabilities are declining, thus reducing the probability that the
agent forgets how to visit a previous "state", as well as by
reintroducing items back into the exploration bonus reward set if their
success probability ever drops too low, thus ensuring that the agent can
always recover. The learning-progress curriculum does not address
derailment as explicitly, but the underlying dynamic exploration reward
does in effect reduce derailment by removing learned items from the
exploration bonus; while the agent is in a state where it does not have
the necessary prerequisites to obtain any of the items still in
exploration bonus reward set, it is incentivized to first obtain the
necessary prerequisites (without taking exploratory actions), before
focusing on exploratory actions again.

\section{Discussion, Conclusion, and Future Work}

We have introduced a learning-progress curriculum with a dynamic
exploration bonus that adapts to the current skill level of the agent.
Experiments were conducted on the Minecraft "Simon Says" tasks. We first
showed that uniform sampling with no exploration bonus performs poorly,
obtaining only 17 tasks and hitting a ceiling at that level where
additional training produces no new learning. We then showed that
combining the main goal-dependent reward with a static goal-independent
exploration bonus increases the number of learned tasks from 17 to 43.
Dynamically adjusting the exploration bonus to only include tasks with
low success probability further increases the number of learned tasks to
70. Sampling tasks using the bidirectional learning progress curriculum
instead of uniform sampling further increases the number of solved tasks
to 82. Moreover, the sampling of tasks elegantly follows the learning
curve of the agent, focusing learning on the frontier of the agent's
skill set as that skill set expands. In addition, the bidirectional
learning progress curriculum, which tracks both tasks with improvements
and deterioration in performance, effectively mitigates catastrophic
forgetting (which we see in the unidirectional learning-progress
curriculum) by resampling tasks that are at risk of being forgotten.

There are various ways in which the learning progress curriculum could
be expanded in future work. First, while the current method was designed
under the assumption that explicit task labels are available, it could
be made more general by developing methods that can dynamically group
tasks into clusters over which learning progress is averaged. Second, if
the number of tasks becomes too large it becomes computationally
expensive to faithfully estimate learning progress for each task. A
promising future direction would be to estimate learning progress over
large tasks spaces using function approximation or, relatedly, generate
environments that are expected to feature high learning progress via a
neural network environment generator.

\begin{ack}
We thank Ilge Akkaya, Bob McGrew, Reiichiro Nakano, Matthias Plappert and John Schulman for discussions, support and feedback on this manuscript.

\end{ack}

{
\small

\bibliographystyle{unsrt}
\bibliography{bibliography.bib}

}

\appendix

\section{Appendix}
\subsection{Optimal $\Delta t$ for estimating learning progress:}
\label{app:a}

Let us model success probability as an i.i.d stochastic process
$x(t)$ with mean $\mu(t)$ and variance
$\sigma^2(t)$. Learning progress is given by the derivative of
the mean, $\mu^\prime(t)$. We want to estimate learning progress
$\mu^\prime(t)$ by averaging $n$ samples from $x(t)$ and
from $x(t - \Delta t)$ and calculating the difference quotient:
The expected square error of this estimator $\hat{\mu^\prime}(t) =
\frac{\frac{1}{n}\sum_i x_i(t) - \frac{1}{n}\sum_i x_i(t-\Delta
t)}{\Delta t}$ is given by:

\begin{equation}
\label{eq:exp_square_error}
    \textrm{err}^2 = \frac{2 \bar{\sigma}^2}{n \Delta t^2} + \frac{1}{4} \mu^{\prime\prime}(t)^2 \Delta t^2 + O(\Delta t^3),
\end{equation}

where $\bar{\sigma}^2 = \frac{1}{2}\left(\sigma^2(t) + \sigma^2(t - \Delta t)\right)$ is the average variance and $\mu^{\prime\prime}(t)$ is the second derivative (curvature) of the mean. The notation $O(\Delta t^3)$ means that we assume $\Delta t$ to be small and that we neglect terms of third order and higher. The curve \eqref{eq:exp_square_error} corresponds to the orange curve in Figure \ref{fig:5}, middle.

The optimal $\Delta t$ that minimizes the square error in \eqref{eq:exp_square_error} is
given by $(\Delta t)_{opt} = \left( \frac{8 \bar{\sigma}^2}{n
\mu^{\prime\prime}(t)^2} \right)^{\frac{1}{4}}$. If we increase the number
of measurements $n$, our estimate of success probability becomes
more accurate and we can afford a smaller $\Delta t$. If the
curvature $\mu^{\prime\prime}(t)$ increases (i.e. the success probability
curve has high curvature), we need to choose a smaller $\Delta t$
to keep the bias of our estimator in check. If the average variance
$\bar{\sigma}^2$ gets larger, we need to increase $\Delta
t$ to keep the variance of our estimator in check. The optimum exists
if and only if $\mu^{\prime\prime}(t)$ is non-zero.

\textbf{Proof of \eqref{eq:exp_square_error}:} 

Since we know the means and variances of the samples of $x(t)$ and
$x(t - \Delta t)$ we can also calculate the mean and variance of
the estimator $\hat{\mu^\prime}(t) = \frac{\frac{1}{n}\sum_i x_i(t) - \frac{1}{n}\sum_i x_i(t-\Delta t)}{\Delta t}$, because the latter
is just defined by a linear combination of $x_i(t)$ and $x_i(t - \Delta t)$.

For the mean we find:

\begin{equation*}
    \left<\hat{\mu^\prime}(t)\right> = \frac{\frac{1}{n}\sum_i\left<x_i(t)\right> - \frac{1}{n}\sum_i\left<x_i(t-\Delta t)\right>}{\Delta t} = \frac{\mu(t) -\mu(t-\Delta t)}{\Delta t}
\end{equation*}

This means that our estimator is an unbiased estimator of the finite
difference quotient, but not of the derivative $\mu^\prime(t)$ (which
we obtain in the limit $\Delta t \rightarrow 0$). We can
calculate the relationship between the two by Taylor-expanding to second
order in $\Delta t$:

\begin{equation*}
    \mu(t - \Delta t) = \mu(t) - \mu^\prime(t) \Delta t  + \frac{1}{2}
\mu^{\prime\prime}(t) \Delta t^2 + O(\Delta t^3)
\end{equation*}

The $O(\Delta t^3)$-notation means that we are neglecting 3rd
and high-order terms in $\Delta t$, because we assume $\Delta
t$ to be small.

Plugging this expression in our expression for the mean yields:

\begin{equation*}
    \left< \hat{\mu^\prime}(t) \right> = \mu^\prime(t) - \frac{1}{2}
\mu^{\prime\prime}(t) \Delta t  + O(\Delta t^2)
\end{equation*}

This expression means that the bias of our estimator is determined (to
first order) by the second derivative of the mean. Note also that the
neglected terms are now of 2nd order and higher, because we have divided
by $\Delta t$.

For the variance we find:

\begin{equation*}
    \textrm{Var}\,\hat{\mu^\prime}(t) = \frac{\frac{1}{n^2}\sum_i \textrm{Var}\,x_i(t) +
\frac{1}{n^2}\sum_i \textrm{Var}\,x_i (t - \Delta t)}{\Delta t^2} =
\frac{\frac{1}{n}\sigma(t) + \frac{1}{n}\sigma(t - \Delta
t)}{\Delta t^2}
\end{equation*}

To arrive at this expression we have made use of several facts and
assumptions: Since all samples from $x(t)$ and $x(t - \Delta
t)$ are uncorrelated we can simply add the variances. Second, if one
multiplies/divides a random variable by a constant, its variance gets
multiplied/divided by the square of the same constant.

Using our expression for the average variance, $\bar{\sigma}^2 = \frac{1}{2} (\sigma^2(t) + \sigma^2(t - \Delta t))$, the variance is given by:

\begin{equation*}
    \textrm{Var}\,\hat{\mu^\prime}(t) = \frac{2 \bar{\sigma}^2}{n \Delta
t^2}
\end{equation*}

The expected square error of our estimator is given by the sum of the square of
the bias and the variance:

\begin{equation*}
    \textrm{err}^2 = \left(\left<\hat{\mu^\prime}(t)\right> -
\mu^\prime(t)\right)^2 + \textrm{Var}\,\hat{\mu^\prime}(t) = \frac{2
\bar{\sigma}^2}{n \Delta t^2} + \frac{1}{4} \mu^{\prime\prime}(t)^2 \Delta
t^2 + O(\Delta t^3),
\end{equation*}

which completes our proof.

\subsection{Reweighting of learning progress towards small success
probabilities}
\label{app:b}

For each task we use EMAs with different time scales to obtain a "fast"
and a "slow" measure of success probability, $p_{\text{fast}}$ and
$p_{\text{slow}}$. Our bidirectional learning progress measure is given
by $LP = |f(p_{\text{fast}}) - f(p_{\text{slow}})|$, where we use the reweighting function

\begin{equation*}
    f(p) = \frac{(1 - p_\theta) p}{p + p_\theta (1 - 2 p)}
\end{equation*}

with parameter $p_\theta = 0.1$.

\begin{figure}[h]
    \centering
    \includegraphics[width=\linewidth]{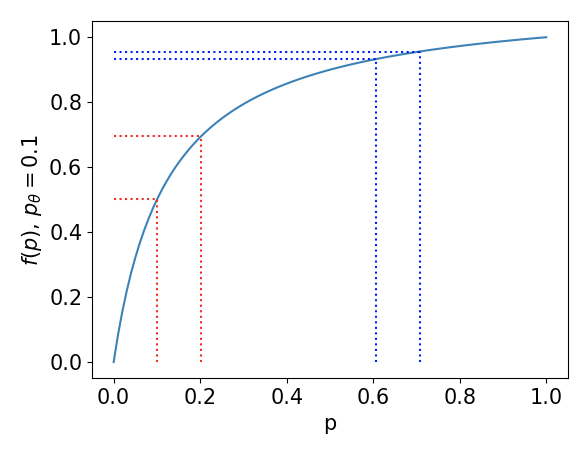}
    \caption{Reweighting of probabilities before calculating learning progress}
    \label{fig:6}
\end{figure}

The reweighting function magnifies differences in small probabilities,
an effect that is illustrated in Figure \ref{fig:6}: A probability
difference between $p=0.1$ and $p=0.2$ (dotted red lines)
leads to a much larger difference in reweighted probabilities than a
probability difference between $p=0.6$ and $p=0.7$ (dotted
blue lines).

\subsection{Conversion of reweighted learning progress measure to sampling
weight}
\label{app:c}

\begin{figure}[h]
    \centering
    \includegraphics[width=\linewidth]{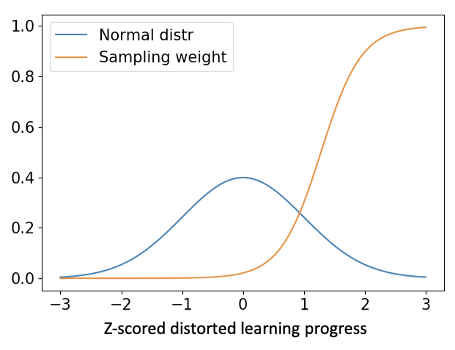}
    \caption{Sigmoid sampling function applied to z-scored reweighted learning progress}
    \label{fig:7}
\end{figure}

When converting reweighted learning progress to sampling weight (i.e.
unnormalized sampling probability) we choose a sampling function that
focuses on roughly 10\% of tasks with the largest reweighted learning
progress, but also prevents overfitting to just one or two tasks. The
algorithm is as follows:
\begin{itemize}
    \item Z-score the reweighted learning progress (subtract mean and divide by standard deviation)
    \item Apply a sigmoid to the result. The sigmoid is centered on 90\% quantile of the normal distribution (Figure \ref{fig:7}). The saturation of the sigmoid for large learning progress prevents sampling from just focusing on one or two tasks.
    \item Normalize resulting weights to sampling probabilities.
\end{itemize}

If the reweighted LP measures were Gaussian-distributed, the above
algorithm would focus on sampling the top 10\% of tasks. In practice the
LP measures often deviate from Gaussianity and we see the curriculum
sometimes focus on a larger or smaller percentage of tasks.

\subsection{Experiment details}
\subsubsection{Environment}
The Minecraft environment the agents interacts with is based on MineRL \cite{guss2019minerl}. At each time step, the agent observes a 64x64 RGB video frame of the player's point-of-view and a set of features about the game-state: player inventory, health, food, saturation, experience point and the item in the main hand and off hand. In addition, the agent observes the id of the current goal item and the items in the exploration set. The main actions of the agent are "attack", "back", "forward", "jump", "left", "right", "sneak", "sprint" and "use" as well as control of the camera. We also provided the agent with special actions to craft, equip and place specific items without using the GUI interface (items that could not be crafted, equipped or placed in a given game state were masked out using an action mask). We used a “frame skip” of 4, i.e. every action by the policy, except “place”, “craft”, “equip” and camera actions, was repeated for 4 consecutive time steps in the MineRL environment before the policy could sample a new action. Given that the Minecraft environment runs at 20 frames per second, this allows the agent to choose a different action every 200 milliseconds, close to the typical human reaction time of 250 milliseconds. Finally, we added a "channel attack" action that had the effect of repeating the "attack" action for a fixed number of policy time steps (the precise number depended on which tool the agent was holding in its main hand at the time the “channel attack” action was taken, with better tools taking fewer steps, because better tools require fewer consecutive attack actions to break a block), which made it easier for the agent to learn how to mine resources.

\subsubsection{Policy and Optimization details}

The policy and value function networks shared the same architecture and weights. Visual observations were processed using an enlarged version of the IMPALA convnet \cite{espeholt2018impala}: the number of filters of the 3 modules were 64, 128 and 128, respectively, instead of 16, 32 and 32 as in the original IMPALA convnet. The convnet was followed by a fully connected layer of size 512. Feature observations were embedded linearly (inventory observations) or using one-hot encoding (all other feature observations) in a 512-dimensional embedding space. We then summed all visual and feature embeddings and processed it through an LSTM layer and another fully connected layer of size 512. The network output was given by a linear, factorial action head for the policy and a linear layer for the value function. Optimization was performed using Proximal Policy Optimization \cite{schulman2017proximal} and General Advantage Estimation \cite{schulman2015high}.

\subsection{Optimization hyperparameters}
\label{app_hyper}

\begin{table}[h!]
\centering
\begin{tabular}{|c|c|}
    \hline
    Buffer size & 209,728 \\
    \hline
    Mini-batch size & 3,264 x 10 time steps \\
    \hline
    Learning rate & $3 \cdot 10^{-5}$ \\
    \hline
    PPO clipping parameter & 0.2 \\
    \hline
    Entropy coefficient & 0.01 \\
    \hline
    $\gamma$ & 0.999 \\
    \hline
    GAE parameter $\lambda$ & 0.95 \\
    \hline
    BPTT truncation length (student) & 10 \\
    \hline
    max episode length & 9000 \\
    \hline
    EMA time scale of learning progress curriculum & 1250 optimization steps \\
    \hline
    1 optimization step & $\sim$ 17180 policy frames \\
    \hline
\end{tabular}
\end{table}

\end{document}